\documentclass[conference]{IEEEtran}
\IEEEoverridecommandlockouts
\usepackage{amsmath,amssymb,amsfonts}
\usepackage{algorithmic}
\usepackage{cleveref}
\usepackage{enumitem}
\usepackage{graphicx}
\usepackage[numbers,sort]{natbib}
\usepackage{textcomp}
\usepackage{todonotes}
\usepackage{xcolor}
\def\BibTeX{{\rm B\kern-.05em{\sc i\kern-.025em b}\kern-.08em
    T\kern-.1667em\lower.7ex\hbox{E}\kern-.125emX}}

\begin{document}

\title{Framework for Certification of AI-Based Systems}

\author{\IEEEauthorblockN{Maxime Gariel}
\IEEEauthorblockA{\textit{Xwing, Inc.} \\
San Francisco, CA USA \\
maxime@xwing.com}
\and
\IEEEauthorblockN{Brian Shimanuki}
\IEEEauthorblockA{\textit{Xwing, Inc.} \\
San Francisco, CA USA \\
brian@xwing.com}
\and
\IEEEauthorblockN{Rob Timpe}
\IEEEauthorblockA{\textit{Xwing, Inc.} \\
San Francisco, CA USA \\
rob@xwing.com}
\and
\IEEEauthorblockN{Evan Wilson}
\IEEEauthorblockA{\textit{Xwing, Inc.} \\
San Francisco, CA USA \\
evan@xwing.com}
}
\maketitle

\begin{abstract}
The current certification process for aerospace software is not adapted to “AI-based” algorithms such as deep neural networks. Unlike traditional aerospace software, the precise parameters optimized during neural network training are as important as (or more than) the code processing the network and they are not directly mathematically understandable. Despite their lack of explainability such algorithms are appealing because for some applications they can exhibit high performance unattainable with any traditional explicit line-by-line software methods.

This paper proposes a framework and principles that could be used to establish certification methods for neural network models for which the current certification processes such as DO-178 cannot be applied. While it is not a magic recipe, it is a set of common sense steps that will allow the applicant and the regulator increase their confidence in the developed software, by demonstrating the capabilities to bring together, trace, and track the requirements, data, software, training process, and test results.
\end{abstract}

\section{Introduction}
Recent technologies including deep neural networks and graphics cards have the potential to enable the automation of functions previously impossible due to algorithmic and computational resource constraints. A common neural networking application is image processing, where cameras have the potential to drastically reduce the size, weight and power of some sensor suites while providing additional functionalities. These benefits arise because a single sensor can be used for multiple purposes (e.g.\ using cameras for airborne intruder detection or lane tracking while taxiing).

Before use on a certified aircraft, all software must undergo a certification effort. The goal of the software certification process is to ensure that the software follows a defined set of requirements and that it does not fail / crash / return error during its execution. Traditionally, this certification process involves linking individual requirements to specific lines of code to check that all requirements are fulfilled and extraneous code is not present. The performance characteristics of the software are evaluated separately. 

For deep neural network (DNN), the precise parameters resulting from training are as important as (or more important than) the code that drives it. Additionally, requirements cannot be tied to specific lines of code because the neural networks functionality is encoded in the large parameter set. We therefore need to provide tools that can track the trained network and the validation data along with the algorithm. Our framework will provide capabilities to bring together, trace, and track the requirements, data, software, training process, and test results. 

While AI will have many applications in aerospace, we use the example of “vision-based detect-and-avoid” to illustrate the proposed framework in this paper. Specifically, we focus on training a DNN for object detection using airplanes. The same reasoning can be applied to other applications like detection of runways or lane tracking for taxiing.

To motivate the need for a new approach, compare a traditional sensor like a radar to a camera for detecting other flying aircraft. Radars have been used for decades and certified for both airborne and ground based applications. 

\begin{figure}[htbp]
\centerline{\includegraphics[width=\linewidth]{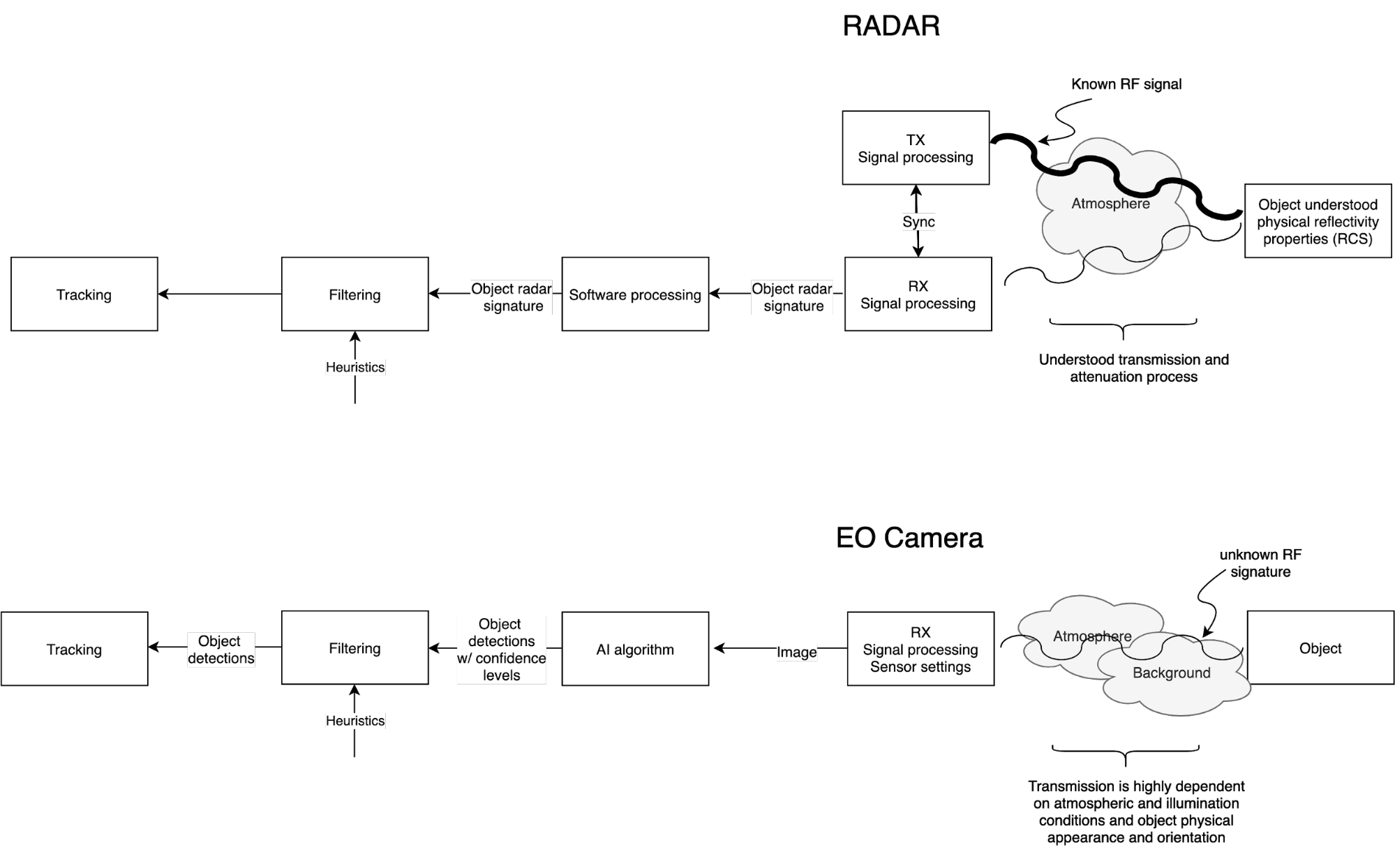}}
\caption{Process for tracking an object using a radar.}
\label{fig:radar}
\end{figure}

As shown in \Cref{fig:radar}, a radar is an active sensor (emits energy) and its physical properties are understood so that every step of the detection can be modeled. Well understood mathematical algorithms including the Fourier transform and the constant false alarm rate algorithm can then be applied to processing radar return data. Additionally, radar data includes time of flight information that is used to determine range which allows for easier discernment of targets from the background.

On the other hand, camera images are more drastically affected by a large range of physical phenomenon like lighting and weather conditions. The large variations in sample data and the challenge of separating targets from background data without range information makes the use of traditional algorithms challenging. Instead, the ``learned description'' captured by a trained DNN can often achieve drastically better results than traditional approaches for image processing tasks. 

In the case of the radar, every element is either known by the manufacturer of the radar or from physics. In the case of vision-based systems, there are a lot of unknowns that cannot be practically modeled, and the software needs to be robust to changes in any of these parameters.

In other words, a radar-based approach could be analyzed using a bottom-up approach, based on a physical model of the world.  A vision based approach, on the other hand, must be analyzed in a top-down manner, where a complex model such as a DNN is trained to perform a task (such as object detection) and its performance is then statistically analyzed to ensure it is adequate.

In particular, we need to ensure that the trained model is generalizable and stable.

\begin{description}
    \item[Generalizability] AI systems are trained with a subset of possible input scenarios but will be used in scenarios across the operational domain. They should generalize to scenarios not encountered during training and still maintain high accuracy.
    \item[Stability] AI systems should behave consistently in similar situations and be robust to small variations. The nonlinear nature of neural networks that allows for great model flexibility also allows for sudden changes in model outputs on images that are nearly identical. The output of the network should be stable and not change suddenly or unexpectedly.
\end{description}

A machine learning model is first trained and then can be used for inference. If the code running the model is deterministic and there is no online learning, then the model is a function from input data to labels/annotations which is determined by the weight parameters that are optimized during training. Thus in two phases, the model is optimized using annotated training data, and then kept fixed and can be run on live data. In the remainder of this paper and framework, we assume that the model is fixed during inference. 

\begin{figure}[htbp]
\centerline{\includegraphics[width=\linewidth]{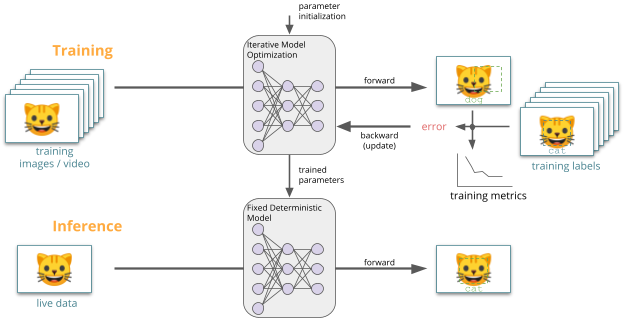}}
\caption{Example of a DNN.}
\label{fig:dnn}
\end{figure}

This two step process of training and then inference is crucial because it allows a thorough analysis to be performed on the trained model with the confidence (at least, in a statistical sense) that the model will perform predictably while in flight.

Once the model is trained, it can be viewed as a black box mapping inputs to outputs. However, because we cannot completely characterize how the weights deciding which additions and multiplications are applied cause certain inferences to arise, it is essential to understand the generalization of the model and where its limitations lie.

When evaluating neural network systems, the input space is too large to test completely. Probability and statistics can be used to show that a system meets performance requirements on a sample set. Thus it is important to ensure that data used to train and test the model spans the operational domain of where the model is to be used.  One of the goals of our framework is to enable this type of analysis to be performed.

The rest of this paper is organized as follows. In Section \ref{sec:related-work} we describe relevant contributions from the fields of machine learning and certification. In Section \ref{sec:framework} we describe the requirements and high-level processes of our proposed framework. In Section \ref{sec:methodology} we describe our work in implementing the core of this framework - a set of tools and processes for tracking aspects of the development process, including datasets, codebases, and trained models. In Section \ref{sec:results} we discuss the results of training and evaluting a DNN for detecting airplanes using this framework. Finally, we give our conclusions in Section \ref{sec:conclusion} and discuss directions for future work.

\section{Related Work}
\label{sec:related-work}

\subsection{Certification Processes}
\begin{figure}[htbp]
\centerline{\includegraphics[width=\linewidth]{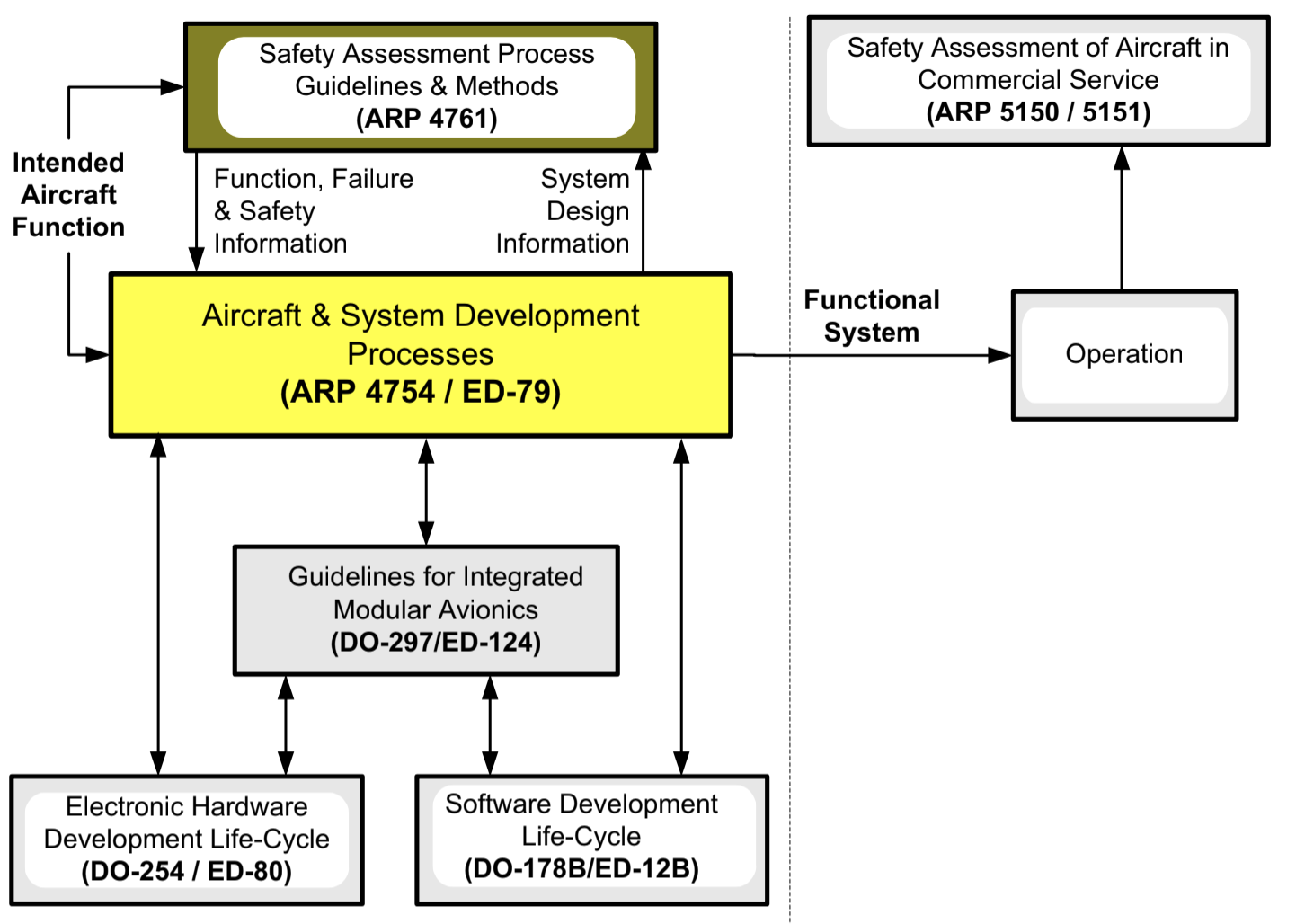}}
\caption{ARP-4754 Process \cite{ARP4754A}.}
\label{fig:arp4754}
\end{figure}

The intent of aircraft certification is to reduce and prevent aircraft accidents. Each set of standards builds to prevent aircraft failures. The widely adopted processes for traditional aerospace system safety and determination of certification levels are defined in ARP-4754 and ARP-4761 \cite{ARP4754A, ARP4761}. Those processes, as illustrated in \Cref{fig:arp4754} guide the manufacturer to determine the level of certification required for hardware and software. DO-254 \cite{DO-254} is the widely adopted hardware certification framework and DO-178 \cite{DO-178C} is the equivalent for software. DO-178 uses the principle that one must trace every line of code to a requirement and vice-versa. That ensures that every line of code is included for a reason and that all requirements are addressed. 

For neural networks, the DO-178 process breaks down when requirements cannot be simply flowed down to explicit lines of code.

EASA released a report \cite{cluzeau2020concepts} of guidelines for learning assurance of machine learning models resulting from a collaboration between EASA and Daedalean in the context of a runway detector project. They recommend verification to be split between data, the learning process, and the inference model. Data should be traceable to its origins and should cover the ConOps space. Learning should be repeatable and convergent. Inference should be verified both in terms of operational guarantees and performance.

\subsection{Machine Learning}
Within the world of machine learning research, there are a number of areas that are relevant to the certification process. Much recent work has been done on the subject of so-called "adversarial attacks". This reflects a concerning property of neural networks, namely that small input perturbations can have a large effect on the output of the network. In adversarial attacks, adversarial examples can be generated which ``trick'' the model into making the wrong classification. These are constructed by making systematic changes to what Ilya et al.\ \cite{ilyas2019adversarial} have called non-robust features. Models can be trained under adversarial training to produce models that are robust to adversarial attacks \cite{madry2019deep}.

While most adversarial attacks have been done directly on the image space, some have looked at more physical realizations. Eykholt et al.\ \cite{eykholt2018robust} have successfully performed attacks by physically perturbing a stop sign before an image is taken. However, for the use case of this paper, at the level of airspace, it is impossible for an adversary to systematically affect large portions of the image formation.

The existence of adversarial inputs has led to an interest in providing formal guarantees that a model is robust to these types of attacks.  For example, Katz et al.\ \cite{Katz_2017} made some progress in formally proving adversarial robustness of DNNs used to approximate the ACAS Xu collision avoidance algorithm.  The authors distinguish between local robustness (i.e.\ the DNN output does not change under adversarial perturbations of an individually specified input) and global robustness (i.e that the DNN output does not change under adversarial perturbations of any possible input).  Their focus is on proving local robustness, while a certifiable system would want to show global robustness, which remains an area of ongoing research \cite{gopinath2018deepsafe}.  Gehr et al.\ \cite{gehr2018ai2} took a different approach, which was better able to scale to the kinds of large DNNs typically used in computer vision. More recent work has focused on proving robustness to adversarial attacks beyond the traditional $L_p$-norm balls \cite{mohapatra2020towards}.

Of course, robustness to adversarial inputs is not the only property needed for certification.  The field of formal verification has made advances in verifying other sorts of properties of DNNs as well.  For example, Reluplex \cite{katz2017reluplex} has been used to formally verify certain properties of DNNs used to represent the ACAS Xu collision avoidance algorithm (such as the property that certain areas of the input space all produce the same output). Yang et al.\ \cite{yang2019correctness} were able to fully verify a neural network in the context of computer vision.  Howevever, their approach requires a full model of the world and the image formation process, which is impractical for our use case (see the example in the Introduction). Unfortunately, current approaches have two major limitations that make practical application in the certification context difficult.  First, these approaches do not scale well to the large networks used in state of the art computer vision (although some recent work has shown promise in this regard \cite{gehr2018ai2} \cite{NEURIPS2018_f2f44698}).  And secondly, many of the properties we might want to verify (i.e.\ the network performs well, regardless of weather conditions) are difficult/impossible to specify formally.  Nevertheless, this is an area of ongoing research and future advances may allow these approaches to be incorporated into our framework.

Another branch of related work is in the area of DNN testing. These methods extrapolate traditional software engineering test methods to apply to DNNs. They automatically generate test cases (i.e.\ synthetic images) in a way that attempts to find edge cases where the model behaves unexpectedly \cite{tian2018deeptest, pei2017deepxplore, zhang2018deeproad, xie2019deephunter}. These test cases can help identify areas where more training data is needed in a much more systematic and thorough manner than manual testing.

This is also a large body of work related to making DNNs more interpretable and explainable \cite{Samek_2021, linardatos2021explainable, Xie2020ExplainableDL}. However, these techniques do not provide the level of interpretability present in traditional software, which leads to the need for the type of framework proposed here.

\section{Framework}
\label{sec:framework}
The objective of the proposed framework is to fill in the gaps where the traditional processes break. We leverage industry standards such as ARP-4754 and ARP-4761 \cite{ARP4754A,ARP4761}, which describe the process of requirements, implementation, and verification. Our framework is not a replacement for this process but rather extends it to provide visibility for neural network systems.

\begin{figure}[htbp]
\centerline{\includegraphics[width=\linewidth]{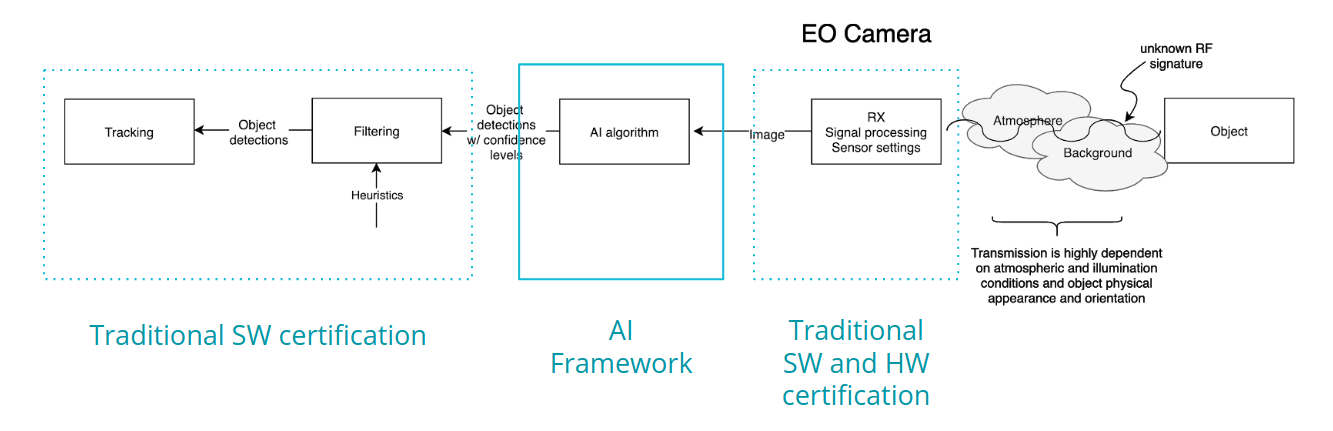}}
\caption{Certification processes for a system with an AI component.}
\label{fig:certification_components}
\end{figure}

As shown in \Cref{fig:certification_components}, a component such as a neural network based object detector needs to be certified alongside more traditional software and hardware components. This framework will serve to demonstrate what needs to be addressed when verifying the neural network component.

\begin{figure}[htbp]
\centerline{\includegraphics[width=\linewidth]{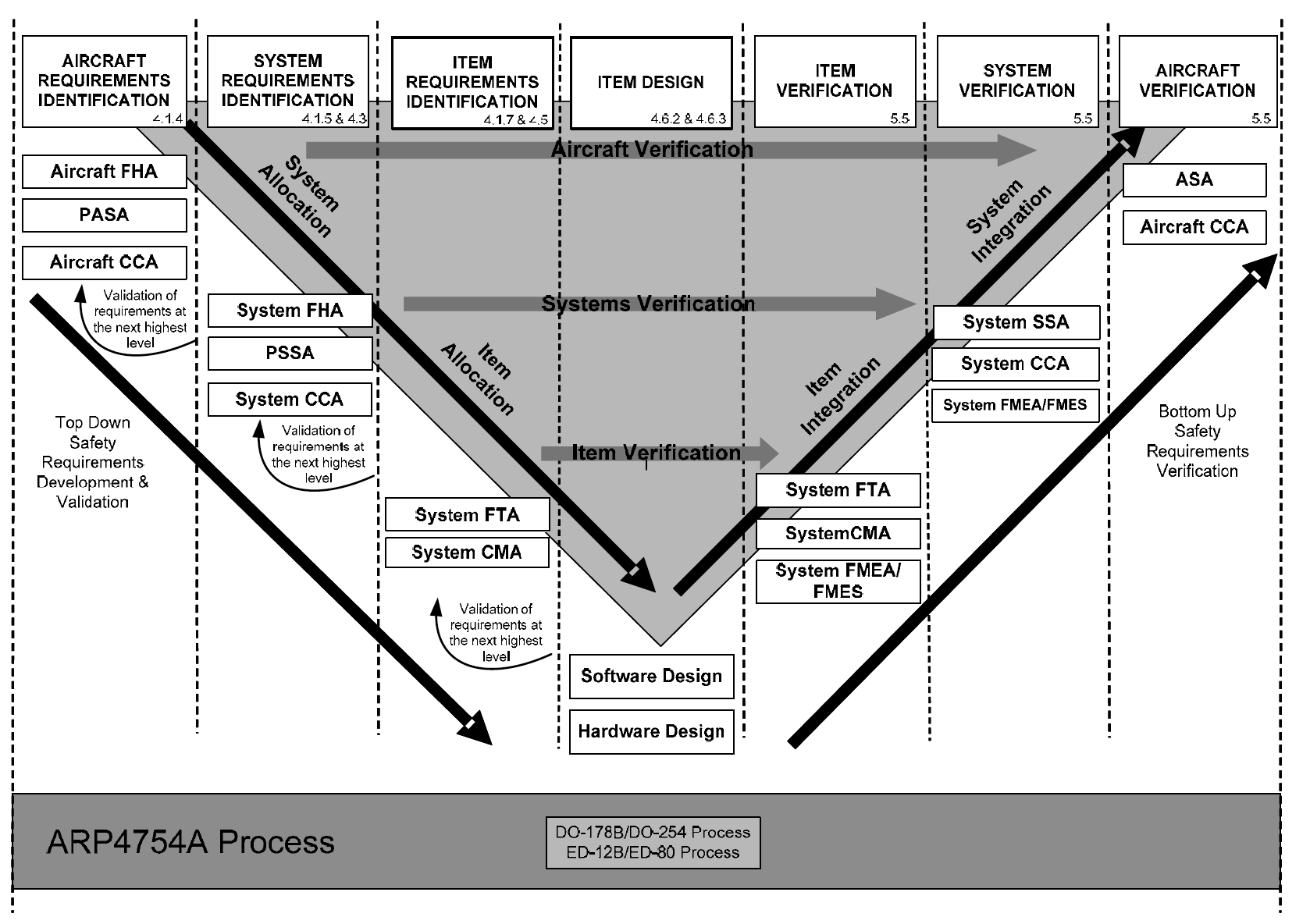}}
\caption{V-model for systems development lifecycle \cite{ARP4754A}.}
\label{fig:vcycle}
\end{figure}

\Cref{fig:vcycle} shows a prototype for the V-model lifecycle from ARP-4574 \cite{ARP4754A}. Requirements are written, for which software and hardware are designed to satisfy, and then are tested and verified. In our framework we will expand on the bottom of the V to capture objectives important for verification for neural networks.

With a neural network, software is written for accelerated hardware (eg. GPUs, TPUs, ASICs, FPGAs) to perform the computations, but the millions to billions of parameters used in the computation are optimized separately and usually stored separately from the committed code base.

A model is a mathematical function performing some objective. In our use case, this is object detection but could be one of a multitude of tasks. The types of models we are considering under our framework are those structured as neural networks.

When a neural network performs a computation, the weights and parameters are just as essential if not more essential than the code running the network computations. The ingredients that affect a model’s behavior are:
\begin{itemize}
    \item The code used to train the model
    \item The initial weights of the model prior to training
    \item The data used to train the model
    \item The code used at inference time
    \item The input image at inference time
\end{itemize}

If the processes for training and inference are deterministic, then the inference model is a function of only these five components, the first four of which do not change at run time.

Our framework assumes a deterministic neural network is used at inference. Online learning is not feasible in our framework as it entails that the model changes while in use and thus inferences are made with an unverified model.

\begin{figure}[htbp]
\centerline{\includegraphics[width=\linewidth]{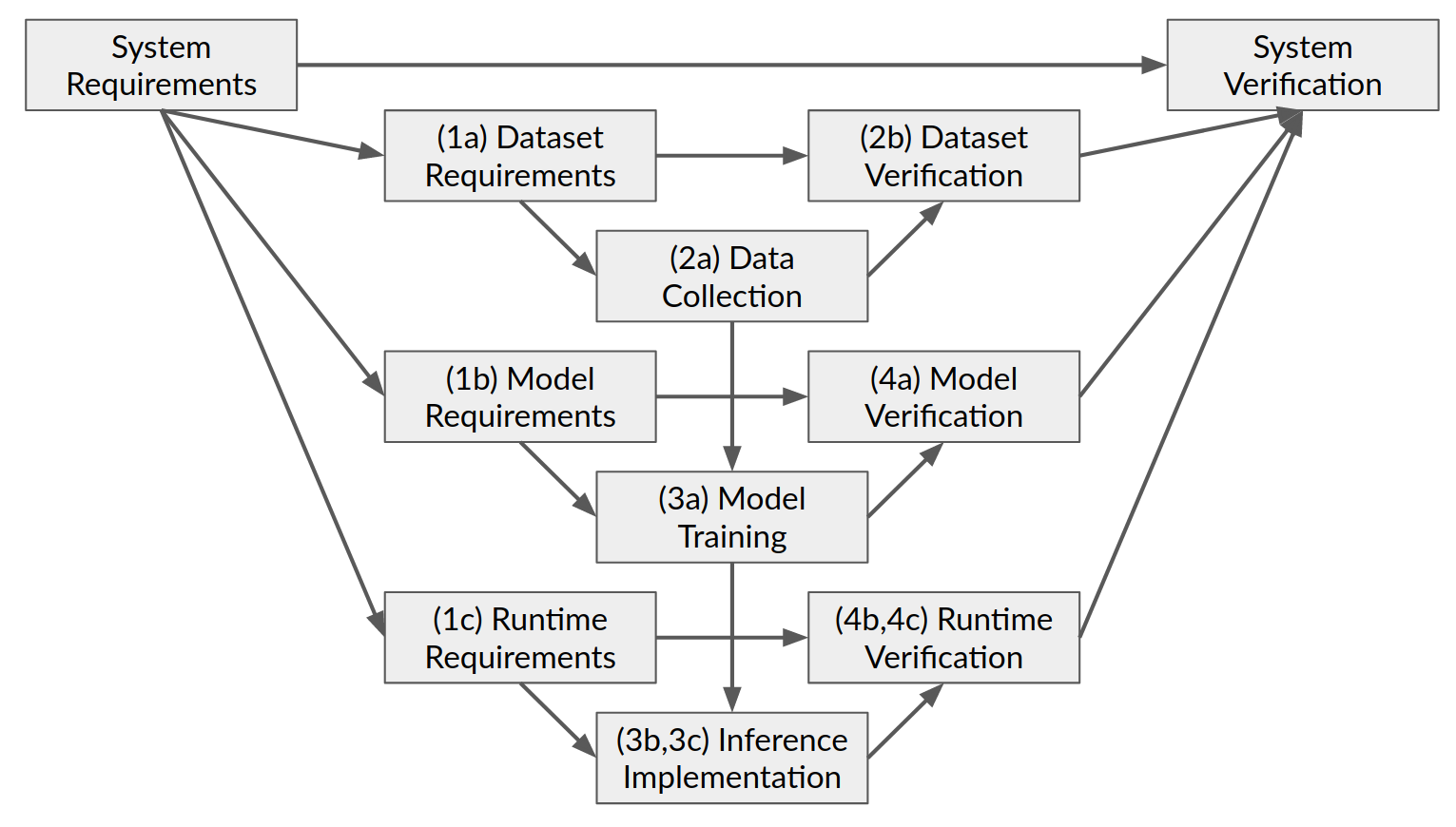}}
\caption{V-model for an AI component, annotated with corresponding framework steps.}
\label{fig:framework_vv}
\end{figure}

Our verification of the model is threefold: we set requirements and processes to verify the data being utilized to build and test the model, the overall model performance on a test dataset, and the reliability of operation at runtime. The dataset verification ensures that this performance spans the types of conditions in the operational domain. The performance metrics on the test set ensure that the model performs well. The runtime verification ensures that the model is executable on the target hardware and environment. We enumerate a series of steps for this in \Cref{fig:steps} and how this fits into a V-model with additional dependencies in \Cref{fig:framework_vv}.

\begin{figure}
    \begin{enumerate}
        \item \textbf{Requirements development:}
        \begin{enumerate}
            \item Define the operational domain
            \item Define model requirements
            \item Define runtime requirements
        \end{enumerate}
        \item \textbf{Data Management:}
        \begin{enumerate}
            \item Data collection
            \item Dataset verification
        \end{enumerate}
        \item \textbf{Implementation:}
        \begin{enumerate}
            \item Model training 
            \item Inference model implementation 
            \item Guardrails/model uncertainty and monitoring
        \end{enumerate}
        \item \textbf{Testing:}
        \begin{enumerate}
            \item Model evaluation
            \begin{enumerate}
                \item Evaluation (i.e.\ performance metrics)
                \item Sensitivity testing
            \end{enumerate}
            \item Runtime verification
            \item Guardrails and Monitoring Evaluation
        \end{enumerate}
    \end{enumerate}
    \caption{Steps in certification framework.}
    \label{fig:steps}
\end{figure}

\subsection{Requirements Development}
Before a model can be evaluated with respect to any use cases, there is a need to identify the operational domain, that is, the conditions in which the system will operate. For our use case, the operational domain should be defined in terms of parameters such as location, environmental/weather conditions, time of day, level of traffic, intruder aircraft attributes, and camera characteristics amongst others. The applicant should make an exhaustive list of all the external conditions that may be encountered, and the proportions in which they are expected to be encountered.
This needs to be in place so that analyses of the data can be measured with respect to the expected set of conditions.

A system is expected to perform a function well and without faults. System analysis is made much more tractable by splitting the requirements into runtime requirements and performance requirements.

The runtime requirements can be verified and validated using traditional methods. For a neural network, these requirements should be about how the system processing the neural network computations functions and cover aspects such as the system's determinism, throughput, latency, and possibility for execution errors.

The performance metrics need to be determined so that the model can be evaluated. In a multiclass detection setting, the mAP (mean average precision) is a commonly used high-level measurement of the model's performance. So a basic requirement might be that the model achieves a certain mAP on the test set, determined by the allowable false negative and false positive rates.

However, the mAP (or any other single statistic, for that matter) is not enough to give a complete picture of the model's performance on its own. Additional, more specific requirements might include:
\begin{itemize}
    \item Maximum overall false positive rate
    \item Maximum overall false negative rate
    \item Accuracy of detections with respect to true target position, typically measured using the intersection over union (IOU) of the detected and ground truth bounding boxes
    \item Maximum false positive/negative rates in specific situations (i.e.\ weather conditions, lighting conditions, intruder distance, etc.).  It may allowable for the model to have degraded performance in rare, less dangerous situations. But it should be possible to show that model performs well, e.g.\ when the intruder is very close, or when visibility is good. It is also expected that during the development process, requirements for additional situations will be added as they are encountered.
\end{itemize}

\subsection{Dataset Collection}
Data must be gathered which spans all of the values of parameters determined in the operational domain, ideally distributed in similar proportions to the distribution of real conditions. This data should be compiled for two types of datasets: the development datasets and the certification dataset. The development datasets, often further split into training and validation sets, comprise the data the applicant uses to construct and train the model. The certification dataset must be kept separate and is used to test the final model.

It would be best if the certification dataset were collected and constructed by an independent party to minimize data dependence, but minimally certification data should be collected on separate flights from data in the development datasets and it should be possible to verify that no images from the certification dataset were used in training.

The data collection process is ongoing throughout the development process. As weaknesses are found in the trained models, new data will need to be collected and added to the training and test sets. Additionally, developers may want to create datasets focusing on specific scenarios, remove images from datasets that are no longer relevant (for example, if different cameras are being used and images taken by older cameras are no longer representative), further refine labels on existing datasets, and otherwise modify datasets.

This creates a need for tools to track changes to a dataset over time. Datasets should be versioned so that it is possible to fully reconstruct any dataset used during the development process. Within each dataset, images should be tracked, along with any relevant metadata, information about the source of the image, and information about the labels in the image. Our approach to maintaining this information is described in Subsection \ref{subsec:version_controlled_datasets}.

\subsection{Dataset Verification}
In order for the performance metrics of the model to be useful for running live inference, the dataset used to test the model needs to be representative of real world conditions, that is, the operational domain. Formally, the following generalization should hold true.

\begin{description}
    \item[Goal] Acceptable performance on the certification dataset implies performance in the operational domain.
\end{description}
To achieve this, certain properties need to hold for the certification dataset.
\begin{enumerate}
    \item  The examples in the certification dataset should cover the variations in operational parameters of the operational domain.
    \item The certification dataset needs to contain examples from throughout the operational domain.
\end{enumerate}

Partitioning by operating parameters ensures that the test conditions vary across the input space. Edge cases should be included to test the boundaries of the model undergoing certification.


The development datasets shall be disjoint from the certification dataset. The trained model needs to be developed on data that is independent from all data in the certification dataset. This ensures that performance on the certification dataset stems from generalization and not directly optimizing for the tested conditions. This means that the data used for the certification dataset should be independently collected from data used for development.

\subsection{Model Management}
A typical development process for a deep learning project will involve repeatedly training models. Each training run may involve different hyperparameters (learning rate, regularization parameters, etc.), model architectures, training code, training data, etc. The output of the training process is a model file, consisting of the neural network architecture and the parameters of the network. These model files should be tracked with the following parameters:
\begin{itemize}
    \item The version of code used to generate and train the model, including all dependencies
    \item Any random seeds and initial weights
    \item Datasets used during training/evaluation
    \item Metrics computed during model evaluation (eg. accuracy, precision/recall, etc)
\end{itemize}

Given this information, we can easily re-train a model using the same architecture, hyperparameters, etc.\ on the same training set to reproduce the same model byte for byte.

\subsection{Inference Model}
The inference model implementation encompasses turning the model into an executable model to be used on the target hardware and environment. If techniques such as batch normalization, dropout, pruning, or quantization are used, the inference model differs from the model in the training process. The metrics above should be computed on the model used for inference rather than the one for training in these cases. Otherwise, it is necessary to verify that the difference between the functions described by the two models are within acceptable bounds.

The system running the inference model also needs to undergo traditional software and hardware verification components such as memory usage and timing. This is easier for models where computations are independent of the data (ie. no branching). There should also be verification that the executable model behaves identically to the model tested during model verification, if there are any differences between the execution environment during evaluation and runtime.

Note that these traditional certification methods do not apply to code used in development (and in particular, code used to train a model) as they deterministically produce model parameters prior to flight. Thus they do not need any runtime guarantees. This code will be tracked and its outputs verified using the processes outlined in our framework.


\subsection{Guardrails and Monitoring}
It is beneficial to detect when a model is not performing as expected and prevent any dangerous behavior that may result. This can be useful during the development process, when unexpected behavior can be addressed by improving the model. However, it is most useful after deployment of a model.

An example of a guardrail to prevent unsafe behavior might be limits on how the model can influence the control systems (i.e.\ to prevent dangerous maneuvers). On the monitoring side, it may be useful to detect when a given input is dissimilar from the training data \cite{haroush2021statistical}. Assertions about the outputs of the model may also be useful, for example to detect flickering (i.e.\ a detected airplane disappearing/reappearing between frames of a video) \cite{Kang2020ModelAF}. However, we leave a full description of this topic out of the scope of this document.

\subsection{Other Extensions to the Framework}
The framework proposed here is largely focused on traceability of datasets and trained models, with the goal of enabling the types of statistical analysis that will be necessary for certification.  And while various types of statistical analysis will certainly be necessary for certification, other techniques may also be useful.

For example, formal verification has the potential to offer much stronger guarantees than any statistical analysis can provide. Challenges remain that prevent formal verification from fully replacing statistical verification in the certification process (see the ``Related Work'' section).  However, as progress is made in this area, formal verification techniques can be increasingly useful. One area of particular focus has been in proving robustness to adversarial inputs. Our framework could be extended by using techniques similar to \cite{gehr2018ai2} in order to show local robustness.  Although this technique can't prove robustness to any possible adversarial attack, it can be used to increase confidence in the model by showing robustness to adversarial examples in the locality of a selection of specific images. Similarly, DeepSafe \cite{gopinath2018deepsafe} has been used to show more general forms of robustness on small networks for low-dimensional input spaces. Future work in this direction may allow for stronger formal guarantees of robustness.

Other sorts of properties that might be formally verified have been less studied and are subject to future research.  For example, future work may allow for an approach similar to that in \cite{yang2019correctness} to be used for verifying the model on a subset of the input space.

Our implementation of this framework provides methods for tracking and analyzing data in the training/test sets, but we do not expect manual dataset creation and analysis to be sufficient for fully covering the input space. Automated testing frameworks directed at DNNs can help in this regard by automatically generating images to test various edge cases of the network \cite{tian2018deeptest, pei2017deepxplore, zhang2018deeproad, xie2019deephunter}. As problematic areas of the input space are discovered and images are added to the training/test sets they will be tracked and analyzed using the process described below.

Automated testing can help find edge cases where the trained model fails, but it may also be useful to understand why a model fails on a given input. This is an area where fields like explainable AI (XAI) may be of use \cite{linardatos2021explainable}. For example, a technique such as LIME \cite{ribeiro2016should} may be useful during the development process to understand why a model fails on a given input image, and what kind of additional training data may be needed to improve the performance. During the certification process it may be helpful to explain the failures of a model on the certification dataset (since perfect performance is unlikely) to better understand the model's failure modes.

\section{Methodology}
\label{sec:methodology}
We have used these concepts in building an object detection model for detecting intruder aircraft in images.

\subsection{Automatic Data Collection and Labelling}
\label{subsec:data_collection}
We collect images with potential intruders automatically during flights and place initial detection bounding boxes using ADS-B and radar data. The quality of these boxes vary as the ADS-B given by other aircraft is not always completely accurate and depending on how well the cameras are calibrated.

These annotations can be refined manually. We track when and how these changes are made.

\subsection{Version Controlled Datasets}
\label{subsec:version_controlled_datasets}
Data is a first-class citizen in our framework. We track data like we track code, by making commits in git.

Data is abstracted into images, annotations, and datasets. Images are the underlying visual data. Annotations are the labels, bounding boxes, and metadata associated with images, either collected automatically or added manually. Datasets are collections of versioned annotations and their associated images. Images are immutable. Annotations and datasets can evolve and update independently. The integrity of all components is guaranteed by tracking the hash. Multiple datasets can coexist, even while pointing at different versions of data items.
We use DVC \cite{dvc} to ensure data integrity of images/videos while keeping the size of the git repository small.

\begin{figure}[htbp]
\centerline{\includegraphics[width=\linewidth]{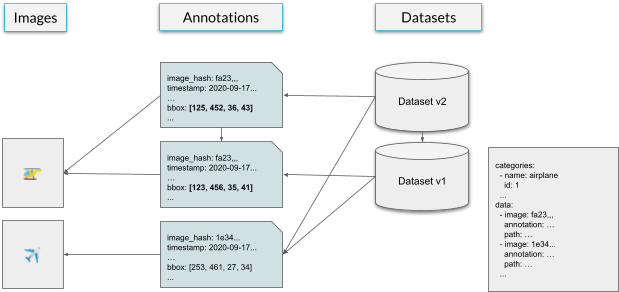}}
\caption{Dataset, annotation, and image traceability.}
\label{fig:datasets}
\end{figure}
In our use case, we refined bounding boxes in annotations using CVAT \cite{cvat}, an open source image and video annotation tool. Updates were committed to our dataset repository, where history is tracked.

\subsection{Training Process}
We train our network with Tensorflow with GPU deterministic operations enabled. All of our batching and image preprocessing is deterministic. When a training job starts, we produce a trace of the versions of all of the code repositories, data repositories, libraries, GPUs, and drivers being used. This way, the training is fully deterministic and repeatable.

\subsection{Data Attributes}
Images extracted from flight data are automatically labeled as described above.  In addition to bounding boxes, data is also labeled with additional attributes that are useful in analyzing the resulting dataset.  These attributes include the callsign of any intruders (which can be used to look up the type of aircraft), time of day, distance to intruder, and position and orientation of the ownship (can be used along with time of day to determine the position of the sun/other landmarks relative to cameras).  Additional information, such as weather/lighting conditions, background type, as well as any other qualitative descriptions of the image, can be added manually.  Together, this information can be used to determine whether a given dataset has an appropriate diversity of images.

Since images are extracted from video, they are also labeled with sequence information.  This makes it possible to analyze the performance of the model from frame to frame, which is important in ensuring stability.

\section{Results}
\label{sec:results}

In order to test and refine our framework, we trained and evaluated a DNN for object detection of airplanes.

\subsection{Datasets}
\begin{figure}[htbp]
\centerline{\includegraphics[width=\linewidth]{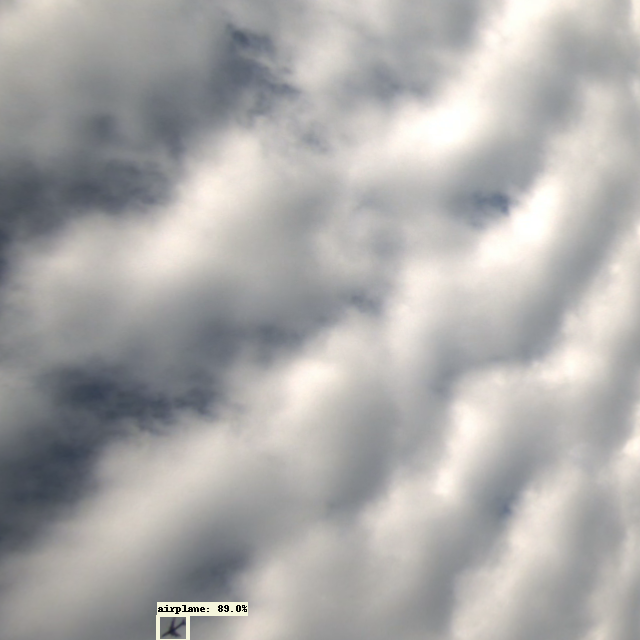}}
\caption{Example aircraft intruder detection.}
\label{fig:detection}
\end{figure}

Images from our flight tests were collected, labeled, and managed using the process described in Subsection \ref{subsec:data_collection}. The labeled data was then split into a training set of 724 images and a test set of 379 images. A typical image with predicted labels drawn is shown in \Cref{fig:detection}. The image has been cropped to a region surrounding the airplane.

\subsection{Performance}
We trained an SSD FPN with ResNet-50 on the training dataset. The model was pretrained on the COCO dataset, and was trained for 50 epochs using the Adam optimizer. Evaluation on the test dataset achieved an average precision of 79\%. For the purposes of this paper we only trained on a single class, the airplane class, and so we use the average precision instead of the more typical mean average precision.

\subsection{Sensitivity by Partitions}
During the certification process it will be important to determine how sensitive a given model is to changes in various properties of the input. For example, a model that loses its accuracy when an intruder aircraft is approaching from a certain angle would need additional training before certification.

To address this, we built a framework for analyzing the sensitivity of a trained model to the types of attributes that we record with our data (see “Data Attributes” section above). To test this approach we performed a basic analysis of the impact of the distance to intruder aircraft on the performance of a trained model. We gathered 379 images from a series of encounters to perform the evaluation. The data was divided into 4 bins based on range - 0-375m, 375-750m, 750-1115m and 1115-1500m.  The average precision was then calculated for each bin. As expected, the model performed better at closer ranges and worse at longer ranges.

\begin{figure}[htbp]
\centerline{\includegraphics[width=\linewidth]{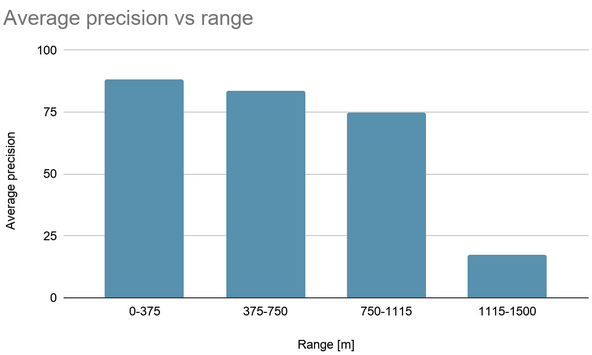}}
\caption{Graph showing the average precision for each range bin.}
\label{fig:AP-by-range}
\end{figure}

\section{Conclusion}
\label{sec:conclusion}
The certification of complex machine learning algorithms presents many new challenges. This project lays out a framework for recommended practices in building and verifying a certified neural network system. In addition, we present some of our work to implement some of that framework by managing the data, training, and verification processes on our use case of detecting aircraft intruders in flight. We illustrate these processes by training a neural network while tracking the origins of how the model was constructed and all data used to train it. This model is not meant to be certified itself, but rather to demonstrate how the types of analysis needed for certification could be performed.

There are many challenges and risks in using machine learning systems in safety-critical applications. By bringing together and tracking the requirements, data, training, and testing, we can evaluate and make statistical guarantees on the domain characterized by the data collected.

The view presented here is a heavily data focused approach for the use of a deterministic vision based system. The ideas of online learning and state-dependent systems (eg recurrent networks) have not been investigated due to additional complexity with little to no additional benefit for our use case. Moreover, we leave other parts of implementing the framework, such as the development of monitoring systems and additional guardrails as realtime checks for future work. Other possible extensions to the framework, such as formal verification, automated testing/image generation, and methods for model explainability/interpretability, are subject to change as these fields advance and are also left for future work.

Our framework focuses on providing methods to ensure that a trained model is generalizable across macroscopic variations in input changes (such as changes in weather, range to target, etc.).  We do not directly address the need to demonstrate stability.  Techniques such as training on adversarial examples can be easily integrated into our framework.  Analyses of model stability can be integrated during the certification phase, for example by generating datasets of perturbed images (adversarial or otherwise) and storing/tracing them using our framework.

\section*{Acknowledgements}
This work was sponsored by the NASA SBIR/STTR program.

\nocite{*}
\bibliographystyle{plainnat}
\bibliography{bibliography}

\end{document}